# Polyploidy and Discontinuous Heredity Effect on Evolutionary Multi-Objective Optimization

Wesam Elshamy, Hassan M. Emara, A. Bahgat

Department of Electrical Power and Machines,
Faculty of Engineering, Cairo University, Egypt

{wesamelshamy, hmrashad, ahmed.bahgat}@ieee.org

**Abstract.** This paper examines the effect of mimicking discontinuous heredity caused by carrying more than one chromosome in some living organisms' cells in Evolutionary Multi-Objective Optimization algorithms. In this representation, the phenotype may not fully reflect the genotype. By doing so we are mimicking living organisms' inheritance mechanism, where traits may be silently carried for many generations to reappear later. Representations with different number of chromosomes in each solution vector are tested on different benchmark problems with high number of decision variables and objectives. A comparison with Non-Dominated Sorting Genetic Algorithm-II [7] is done on all problems.

## 1. Introduction

Evolutionary Multi-Objective Optimization (EMOO) algorithms are evolutionary in that they evolve with the problem. Their power is in their evolving implicit rules. Whenever we impose deterministic static rules on them, we detract from their power. These imposed rules make the algorithm less evolutionary and more deterministic. It even fails to achieve the explicit goal it was created for when the same goal takes another form.

By enforcing an explicit procedure on an Evolutionary Algorithm (EA) to maintain good diversity of solutions, we are only paving paradise to look better. This procedure which explicitly favors some solutions over others to maintain good diversity restricts the algorithm from evolving freely. It may not provisionally accept poor distribution to escape local minima. The algorithm becomes less evolutionary. The notion of accepting worse solutions to overcome a hurdle should not be limited to convergence in EMOO. In fact it should not be limited to any performance criterion if we want the algorithm to be more Evolutionary Algorithm. One way to do so is to mimic natural evolution by mimicking living organisms' building blocks and their environment.

Genetic Algorithms (GA) [3,8,9] was invented amid our continuous effort to mimic nature to solve our problems. It provides the building blocks (genes' representations) and the environment (selection, mating and survival rules). We need to expand this set of building blocks and provide a more evolutionary environment. Researchers, instead, were confined to a limited set of building blocks and introduced some of their well studied deterministic methods to the environment.

Though nature provided us with varied representations for living organisms' genes, most researchers built their models on the first GA models [3,8,9] due to their good results [2]. The effect of carrying redundant chromosomes in EA representation [1,2,3,9] remains under-researched. In nature, polyploidy which is having more than one set of identical chromosomes in each cell is found in many plants. Some plants such as wheat developed through millennia of hybridization strands that have two, four and six sets of chromosomes. Each of these strands was developed to adapt to certain environmental conditions.

In nature, polyploidy helps in maintaining diversity among species. It may even produce new species. This may happen naturally or induced artificially to produce species with desired characteristics. In plant breeding, the induction of polyploids is a common technique to overcome the sterility of a hybrid species [6]. Polyploidy helps organisms adapt to their environment. In polyploid organisms, some alleles are kept in abeyance and termed to be recessive under certain environmental conditions. These conditions reward the emergence of other alleles known to be the dominant alleles. Whenever these conditions change so does the dominance relationship changes.

Though there are dominance schemes other than simple dominance, such as partial dominance and co-dominance, their effects were not investigated before. This is mainly because most research on polyploidy was carried out on binary problems. In partial dominance, an intermediate value between parents' alleles is expressed. It provides even more population diversity than simple dominance in which a distinct parent allele is expressed. The model we present in this paper differs from other polyploid models in that; i) Uniform crossover of each parent chromosomes precedes recombination. ii) New alleles for the offspring are created from parents' alleles using partial dominance.

The organization of this paper is as follows. In part two we present a biological background about genetics and Mendelian inheritance and introduce some genetic related terms. While in part three we review previous related work that investigated the effect of polyploidy in EA. Then we introduce the representation of solution vectors and the variation operators used in part four. In part five we investigate the effect of different ploidy on convergence and diversity and compare it with Non-Dominated Sorting Genetic Algorithm II (NSGA-II). We conclude in part six by conclusions and proposed future research.

## 2. Biological Background

Genes are the blueprints of our bodies. They do instruct cells how to function. Genes are part of chromosomes which are long strand of chemical substances called *deoxyribonucleic* acid (DNA). Humans have 23 pairs of such chromosomes with estimated 30,000 genes in each single set of these chromosomes. So we have two copies of these 30,000 genes. Though a set of unique chromosomes may contain many chromosomes (23 chromosomes in humans), we will simply refer to this set of chromosomes as a chromosome in the remainder of this paper for convenience, unless otherwise stated. Each gene has a value and the values of our genes, which are known as the *genotype,* determine our characteristics, known as the *phenotype,* from hair color to voice pitch and body size. Each different value of a gene is called an allele, so there are blue and brown eye color alleles and they are different versions of the eye color gene. A gene may determine one characteristic or more, or act with other genes to shape a characteristic (this is known as *epistasis*).

During matting, a process of meiosis in which recombination of each parent chromosomes is done. Then they get split in each parent to produce the haploid (contains half the number of chromosomes in a cell) gamete cells, which is equivalent to the male sperm and female egg. Then these gametes fuse together producing the new offspring complete cells containing twice as much chromosomes as each gamete cell [6]. Since a child carries both of his parents alleles, his expressed characteristics are determined according to the dominance scheme. In simple dominance scheme, his characteristics are determined by the dominant alleles in either homozygous (have two copies of the same allele) or heterozygous (have different alleles for a certain gene) cells, or by the recessive alleles of the homozygous cells. The following example explains Mendelian inheritance in rabbits, which assumes a simple inheritance scheme.

In this inheritance scheme, the rabbits are diploid because each one has two chromosomes in his cells (one from each parent). Each one of these chromosomes may have a different color allele. In the reproduction process the two chromosomes are crossed-over and separated to make different combinations with the other two chromosomes from the other parent. In Figure 1, the letters under rabbits indicate their color alleles. A capital letter "**B**" indicates the dominant color allele which is white, while the small letter "**b**" is the recessive color allele, which is black. Two homozygous parents bread four rabbits. By recombining the parents color alleles we get the offspring color alleles as shown in Figure 1. All four rabbits of the first generation are white due to the dominant white color allele they inherited. But in the second generation produced by two rabbits from the first generation, a black rabbit emerges due to the absence of the dominant white color allele in his chromosomes.

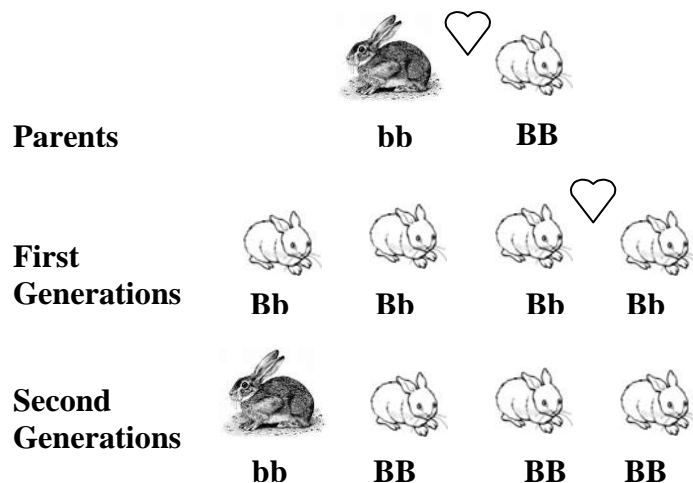

Figure 1 Mendelian inheritance

This shows how some alleles are held in abeyance when they are recessive (black allele) and a full generation may be produced without any expression of the recessive alleles. Some species have more than two chromosomes in their cells, they are called polyploidy. They are very rare among humans (happens mainly due to mutations and they do not survive), and limited in animals, but found in many plants [6]. Some plants such as wheat developed strands that are diploid (two chromosome sets in a cell), tetraploid (four chromosome sets) which is known as macaroni wheat and hexaploid (six chromosome sets) known as bread wheat. It is still unclear why some species has higher number of unique chromosomes in each cell than others and why some has more copies of these chromosomes.

Other dominance schemes include partial dominance and co-dominance. In partial dominance (also known as incomplete dominance) more than one allele affects the phenotype. A classical example of partial dominance is the color of the carnation flower that takes variants of the red color due to the presence or absence of the red pigment allele. In co-dominance scheme, both alleles are expressed. A well known example for co-dominance is the Landsteiner blood types. In this example, both **A** blood type alleles and **B** blood type alleles are expressed leading to an AB blood type carrying both phenotypes.

## 3. Related Work

Early work examining the effect of ploidy in GA goes back to 1967 in Bagley's dissertation [1] as he examined the effect of diploid representation. In his work he used a variable dominance map encoded in the chromosome. A drawback in his model was the premature convergence of dominance values which led to an arbitrary tie breaking mechanism [2]. This work was followed by a tri-allelic dominance scheme used by Hollstien [9] and Holland [3]. They added for each allele a dominance value associated and evolving with it. It took values of 0, a recessive 1 or a dominant 1, though they used different symbols.

Unlike pre-mentioned works which was done on stationary environments, Goldberg and Smith [2] experimented on non-stationary environment. They used an 0-1 knapsack problem with two evolving limiting weights. They concluded that the power of multiploidy is in non-stationary problems, because of the abeyance of the recessive alleles that remembers past experiments. But they didn't show the performance of their algorithm in remembering more than two oscillating objectives. This was a big shortcoming, because most real world non-stationary problems are non-cycling problems. They may come out of order, and sometimes never repeat. Ng and Wong [11] argued that the enhanced performance in [32] was due to the slow convergence encountered in the diploid representation. They proposed a dominance scheme that used dominant (0, 1) and recessive (0, 1) alleles and inverted the dominance of alleles whenever the individual's fitness fall below a 20% threshold value. They reported an enhanced performance over tri-allelic representation. Some researchers extended the application of polyploidy beyond GA. Polyploidy was applied to Genetic Programming (GP) as well [12]. In his comparison of these algorithms Branke [10] Notes:

> Given the evidence available so far, it can be assumed that the multiploid representations may be useful in periodically changing environments where it is sufficient to remember a few states and where it is important to be able to return to previous states quickly. The applicability to problems without periodicity and more than a few re-occurring states is at least questionable.

We note that all these researches were conducted for single objective optimization problems. Most of them used relatively low number of decision variables and binary problems such as 0-1 knapsack problem. At the same time many of these investigations were concerned with manipulating the simple dominance scheme and comparing these variants. The monoploid number (number of chromosomes in each solution vector) remained constant in most of these investigations. Scarce applications such as [13] were done on Multi-Objective Optimization Problems (MOOP). In [13] they used diploid vectors to search in 2-dimensional space optimizing 3 objectives for a food extrusion process. First they worked on each objective separately and produced offspring better than the worst individual for this objective (otherwise the offspring is killed). Then Pareto dominance is applied to the combined populations of each objective.

What is common among all previous work is that they used simple dominance in which either the allele dominates or recesses, which is not always the case in nature. Partial dominance and co-dominance were not investigated before. Partial dominance produces new phenotypes that help the population to adapt to totally new environments and to remember them. It provides more phenotype diversity than simple dominance.

## 4. Representations and Algorithm Procedure

In biology, ploidy is the number of sets of chromosomes in a cell. So a cell that has one set of chromosomes is a monoploid cell, while the one that has two sets is a diploid cell (humans' cells are diploid, except gamete cells), three sets make it triploid, and so on. We use the term *"d-ploid"* to indicate the ploidy number, so 1-ploid representation is a monoploid one, and 2-ploid representation is a diploid one, and so on. In our algorithm we use the partial dominance scheme. The phenotype of each solution vector is based on the partial dominant alleles, or the Dominant-Alleles-Set (DAS), shown in Figure 2. The DAS determines the fitness of each solution vector, and its expressed decision variables values as well. The ploidy number $d$ of a solution vector is the number of all chromosomes in that vector, including the DAS. In the mating process; first, for each parent, one allele representing each locus of a chromosome is chosen by random from all chromosomes in that parent creating one set of alleles. This is analogous in biology to the crossover between each parent chromosomes before they split. Then partial dominance is applied to these two sets of alleles producing the child's DAS. After that, *d-1* chromosomes are randomly chosen from both parents to fill the remaining chromosomes in the *d*-ploid offspring vector. Mutation operator is applied to the child's DAS, while the other *d–1* chromosomes are not mutated.
Figure 2 explains this mating procedure. A detailed procedure of the algorithm is as follows;

**Initialization:** for a *d*-ploid representation. A population of size *n* is initialized at random by filling each DAS and the *d-1* chromosomes for each of the *n* decision vectors. Then the fitness of each decision vector is evaluated according to its DAS.

The domination of solution vectors is determined based on the regular Pareto dominance.

**Mating Selection:** only non-dominated solution vectors are selected for mating. This will result in a varying mating pool size and consequently a varying offspring size for each generation. This causes a very high selection pressure. It will be shown later that the redundant chromosomes will help the population maintain a good diversity of solutions and inhibit premature convergence.

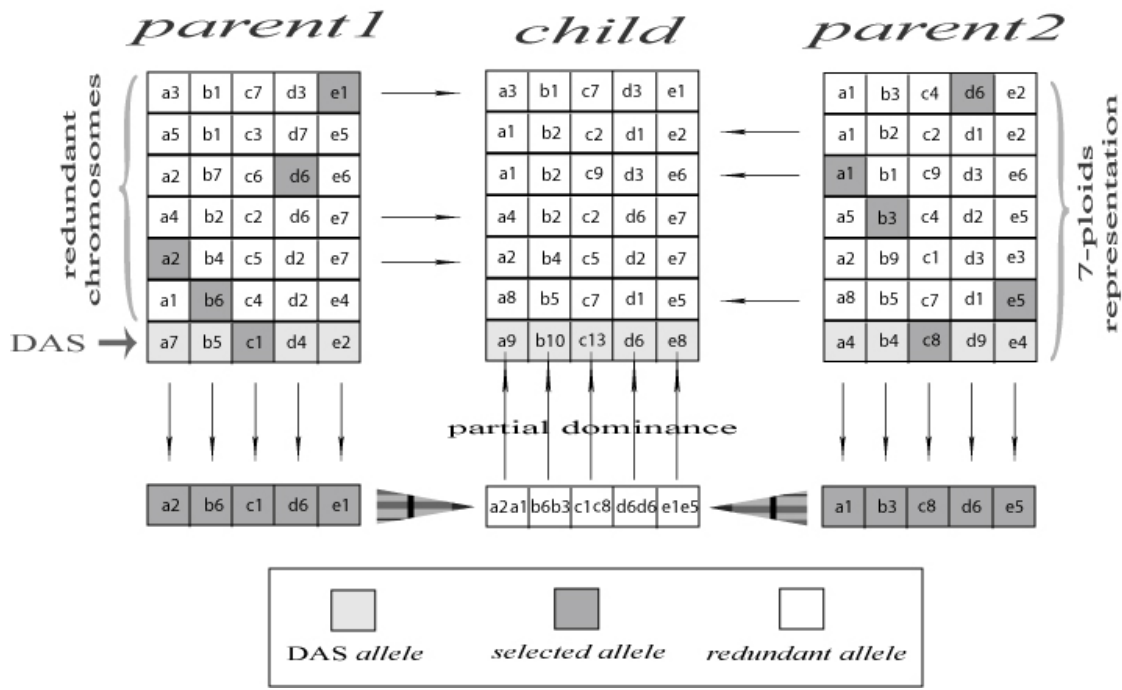

Figure 2  Mating procedure

**Variation Operators:** after filling the varying size mating pool, two parents are selected at random from the mating pool and each individual is allowed to mate only once. After their selection, for each parent, an allele representing each locus of a chromosome is selected by random from all available alleles in this locus. These alleles create two sets of alleles, one for each parent. Any recombination operator may be applied to those two sets to create the child's DAS. In our experiment we use Simulated Binary Crossover (SBX) [14]. Mutation is then applied to the child's DAS. Any mutation operator may be applied here, so we use polynomial mutation [14]. After the child's DAS is created, the remaining *d-1* chromosomes of this child's solution vector are selected and copied at random from both parents' chromosomes. All the chromosomes in both parents have the same probability of being selected and copied to affect later generations. No mutation is applied on these redundant *d-1* chromosomes.

**Survival Selection:** after evaluating the offspring's fitness. Parents and offspring fight for survival as Pareto dominance is applied to the combined population of parents and offspring. Then the least dominated *n* solution vectors survive to make the population of the next generation. Ties are resolved at random.

## 5. Experiments and Results

In the following experiments we use two running metrics to understand the behavior of the algorithms. The average orthogonal distance of solution vectors to the true Pareto front is used to measure convergence because the equations of the global front are known in advance. While the diversity is measured using a modified version of *diversity metric2* [4]. The modified *diversity metric2* and its effect are shown in Figure 3 for a DTLZ2 [15] problem and explained as follows:

For each objective, we calculate a diversity value as follows;

i) For a given objective, the obtained Pareto front using a population of size $n$ is divided uniformly creating $n$ equal cells on the front surface, such as cells $a$, $b$ and $c$ shown in Figure 3. The projection of these cells on the current objective dimension gives unequal $n$ small grids. Then the obtained solutions are projected as well on this objective dimension.

ii) For every projected cell that contains one or more projected solution, an occupation value of **1** is assigned to it, a value of **0** is assigned otherwise.

iii) A diversity value is assigned to each non-boundary cell using a sliding window according the values shown in Table 1, where the cell index is $n$. And the diversity values for boundary cells are assigned according to Table 2, where the boundary index is $k$ for the left boundary case, and is $k+1$ for the right boundary case.

iv) The cells' diversity values are added and divided by the number of cells giving the current objective's diversity value in the range [0, 1].

v) The previous steps are repeated for the remaining objectives. The average of diversity values obtained for all objectives is the overall diversity value for the population. The best population distribution yields a diversity value of **1** while **0** is the worst distribution.

From Figure 3 we can see how the unmodified *diversity metric2* is unable to detect the gap $a$ in the obtained Pareto front when evaluating the diversity for objective $f_1$, and falsely detected a gap close to the boundary $f_1=0$. It is unable to detect either gap $a$ or gap $c$ too in the diversity evaluation for $f_2$, and again falsely detected a gap close the boundary $f_2=0$. This shortcoming in the unmodified version is due to its poor cells partitioning in each objective, because it ignores the shape of the Pareto front. As a consequence, regions with higher slope are less represented by reference points, and so gaps are overlooked in these regions. On the other hand, regions with lower slope get higher than needed reference points leading to false gaps detection. Our modification exploits the knowledge of the Pareto front in benchmark problems and changes the projected cell size to reflect the shape of the Pareto front. It provides more accurate diversity measure for benchmark problems.

In our experiments we use some of the DTLZ benchmark problems set [15] to investigate the effect of polyploidy. We use high number of variables to make the problems hard for the algorithms in order to magnify the difference between results obtained from different algorithms.

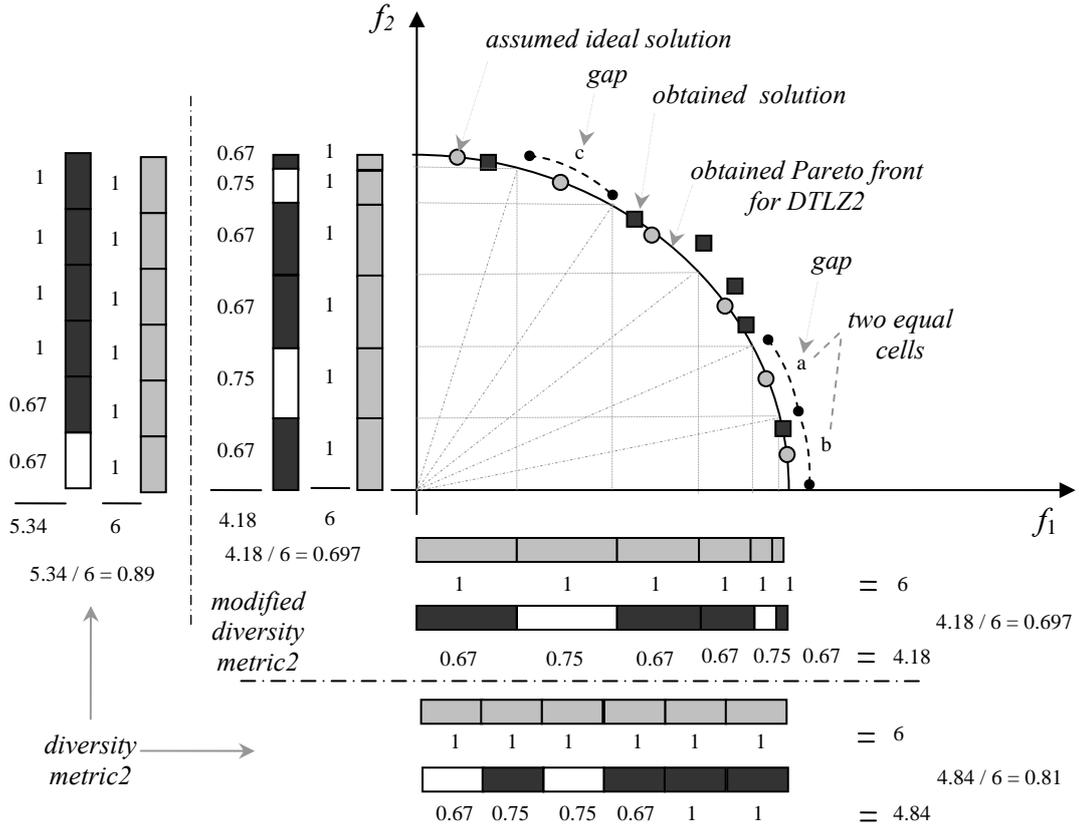

Figure 3: modified *diversity metric2* for DTLZ2

| occupation values | | | cell diversity values |
|---|---|---|---|
| cell ($n$-1) | cell ($n$) | cell ($n$+1) | |
| 0 | 0 | 0 | 0 |
| 0 | 0 | 1 | 0.5 |
| 0 | 1 | 0 | 0.75 |
| 0 | 1 | 1 | 0.67 |
| 1 | 0 | 0 | 0.5 |
| 1 | 0 | 1 | 0.75 |
| 1 | 1 | 0 | 0.67 |
| 1 | 1 | 1 | 1 |

Table 1: non-boundary-cell diversity values

| occupation values | | cell diversity values |
|---|---|---|
| cell($k$) | cell($k$+1) | |
| 0 | 0 | 0 |
| 0 | 1 | 0.67 |
| 1 | 0 | 0.67 |
| 1 | 1 | 1 |

Table 2: boundary-cell diversity values

### *5.1. DTLZ2*

The first test problem we use is DTLZ2 with a high number of variables; $n = 40$, and with 3, 4, 6 and 10 objectives. We use SBX for recombination (with $p_c = 1$ and $\eta_c = 20$), and use polynomial mutation with ($p_m = 1/n$ and $\eta_m = 15$). We investigate the polyploidy effect on convergence and diversity. The results show significant improvement regarding convergence. Figure 4 shows the average distance of each algorithm to the true Pareto front against function evaluations. We can see that the 2-ploids algorithm achieved the best convergence results in the shown cases. We notice that with increasing the number of objectives, the performance of higher ploidy representation and lower ploidy representation are getting close. In the case of 10 objectives, the 10-ploids representation overcomes the 7-ploids representation after around 10,000 evaluations.

It is clear from Figure 4 **a**, **b** and **c** how NSGA-II convergence performance is deteriorating by increasing the number of objectives. It totally lost its way and is diverging in the 10 objectives problem and the ploids algorithms are performing much better. Table 3 shows the diversity of the different algorithms. We can see that the performance of NSGA-II is declining with increasing the number of objectives. On the other hand, the performance of the ploids algorithms is either steady or improving with increasing the number of objectives. In the 3 objectives case, NSGA-II has a superior diversity value of 0.799, while the 7-ploids representation achieves a lower diversity value of 0.6452. But for the 10 objectives problem, the 7-ploids algorithm overcomes NSGA-II by achieving a value of 0.7398 while NSGA-II achieves 0.7194.

To analyze the performance of the redundant chromosomes we do the following. We create a new population by extracting all the chromosomes in each solution vector. This new population has a size of $n \times d$, where $n$ is the original population size and $d$ is the ploidy number (total number of chromosomes in each solution vector). We calculate the average distance of the new population to the true Pareto front, and evaluate the percentage of the dominated solutions in the new population. For each of the ploids algorithms in Table 4, the first row shows the average distance of the original population to the true Pareto front. The average distance of the new population to the true Pareto front is in the second row. The third row shows the percentage of dominated solutions in the new population.

We notice that the average distance of the new population is slightly worse than that of the original population. In the case of 2-ploids with 3 objectives, the average distance is 0.0515 for the original population and it is 0.0518 for the new population which is only worse by 0.58% than the original population average distance. But in the case of 10-ploids with 3 objectives, the average distance deteriorates from 0.34 to 0.5, which is about 47% deterioration. We notice too that the percentage of dominated solutions is low in the case of 2-ploids with a maximum value of 7.52% for the 3 objectives case. This value is steadily increasing with increasing the ploids number in all objectives cases reaching 70.36% dominated solutions for the 10-ploids with 3 objectives.

The new population which offers $n(d-1)$ more solutions may be used instead of the old population for the 2-ploids algorithm giving the decision maker more choices. It is best used in problems with high cost of function evaluations, as the $n(d-1)$ extra solutions are produced without any extra evaluations.

Figure 5.**a, b** show how the 2-ploids algorithm converged well to the Pareto front. It achieved a reasonable degree of diversity considering the high number of decision variables used ($n$=40). While Figure 5.**c** shows how the front obtained by NSGA-II is far from the true Pareto front

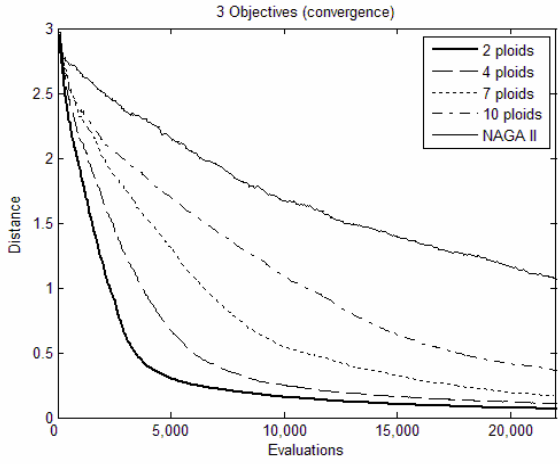

a. 3 objectives

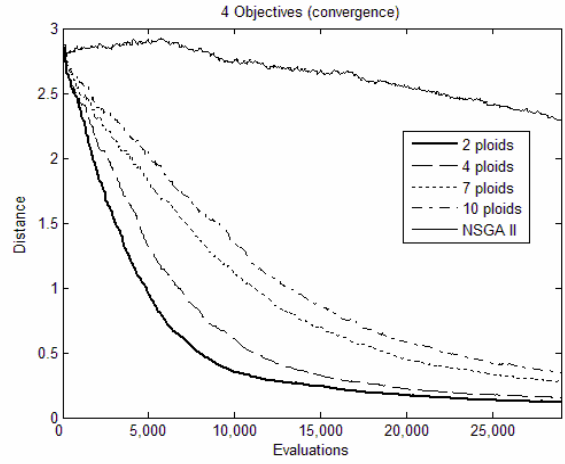

b. 4 objectives

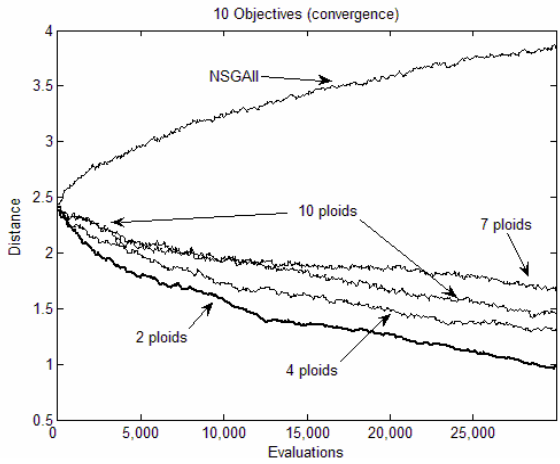

c. 10 objectives

Figure 4 The effect of varying ploidy on convergence speed for DTLZ2 ($n = 40$)

Table 3 The effect of varying ploidy on diversity for DTLZ2 ($n = 40$)

| MOEA | | Objectives | | | |
|---|---|---|---|---|---|
| | | 3 | 4 | 6 | 10 |
| 2-ploids | Average | 0.6621 | 0.6594 | 0.6263 | 0.6872 |
| | Std. Dev. | 0.0208 | 0.0744 | 0.0543 | 0.0721 |
| 4-ploids | Average | 0.6622 | 0.6539 | 0.661 | 0.7198 |
| | Std. Dev. | 0.0114 | 0.0212 | 0.0416 | 0.0492 |
| 7-ploids | Average | 0.6452 | 0.6807 | 0.6773 | 0.7398 |
| | Std. Dev. | 0.0432 | 0.0229 | 0.0175 | 0.0115 |
| 10-ploids | Average | 0.5865 | 0.6459 | 0.6881 | 0.7153 |
| | Std. Dev. | 0.0449 | 0.0229 | 0.0232 | 0.0338 |
| NSGA-II | Average | 0.7997 | 0.775 | 0.7479 | 0.7194 |
| | Std. Dev. | 0.0148 | 0.0159 | 0.0109 | 0.0082 |

| MOEA | Distance to front | No. of Objectives | | | |
|---|---|---|---|---|---|
| | | 3 | 4 | 6 | 10 |
| 2-ploids | Original pop. | 0.0515 | 0.1139 | 0.4381 | 0.9738 |
| | New pop. | 0.0518 | 0.1145 | 0.4428 | 0.9929 |
| | % dominated | 7.52 | 6.25 | 4.89 | 2.11 |
| 4-ploids | Original pop. | 0.0871 | 0.1469 | 0.7348 | 1.3115 |
| | New pop. | 0.089 | 0.1499 | 0.7528 | 1.3463 |
| | % dominated | 19.64 | 13.57 | 10.31 | 3.105 |
| 7-ploids | Original pop. | 0.1402 | 0.2692 | 0.9319 | 1.689 |
| | New pop. | 0.1484 | 0.2785 | 0.9919 | 1.7293 |
| | % dominated | 40.173 | 20.55 | 13.43 | 3.264 |
| 10-ploids | Original pop. | 0.3402 | 0.3358 | 1.2231 | 1.4619 |
| | New pop. | 0.5006 | 0.3685 | 1.316 | 1.4983 |
| | % dominated | 70.36 | 36.67 | 18.52 | 5.08 |

Table 4 performance of the extracted population

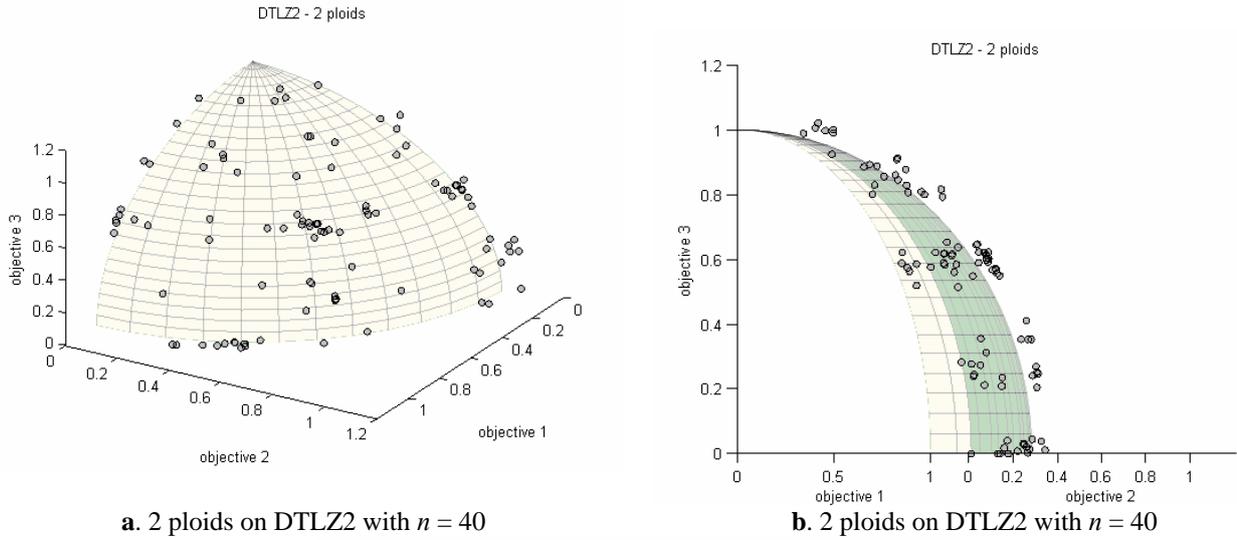

**a**. 2 ploids on DTLZ2 with $n = 40$  **b**. 2 ploids on DTLZ2 with $n = 40$

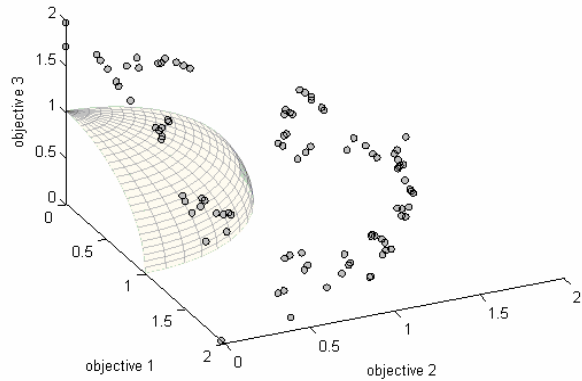

**c**. NSGA-II on DTLZ2 with $n = 40$
Figure 5

### *5.2. DTLZ1*

The second benchmark problem we test the algorithms on is DTLZ1. The difficulty of this problem is in its huge number of local optima. There exists ($11^{(n - M + 1)}$ - 1) local fronts, where *n*, *M* are the number of variables and the number of objectives respectively. For this test problem, we use $n = 30$, with 3, 4, 6 and 10 objectives. The recombination and mutation parameters are the same as those used in DTLZ2. Such high number of decision variables we use creates a huge number of local optima to test the ability of the different algorithms to escape it. We let the algorithms to go for 50,000 function evaluations to analyze the performance in long runs.

From Figure 6 we can see that NSGA-II has the best convergence speed in early evaluations. But it get overcome by the ploids algorithms one after the other, except for the 10-ploids in the 3 objectives case which does not catch it in the scope of our 50,000 function evaluations. We notice too that by increasing the number of objectives, the ploids algorithms catch NSGA-II in earlier function evaluations. In the 3 objectives case, the 7-ploids algorithm catches NSGA-II after around 35,000 evaluations, while the 10-ploids algorithm is unable to catch it. But in the 4 objectives case, the 7-ploids algorithm overcomes NSGA-II after around 23,000 evaluations while the 10-ploids algorithm does so after 33,000 evaluations. This reveals the ability of the ploids algorithms to handle many objectives in highly rugged objective functions. This ability is magnified when the problem gets more difficult.

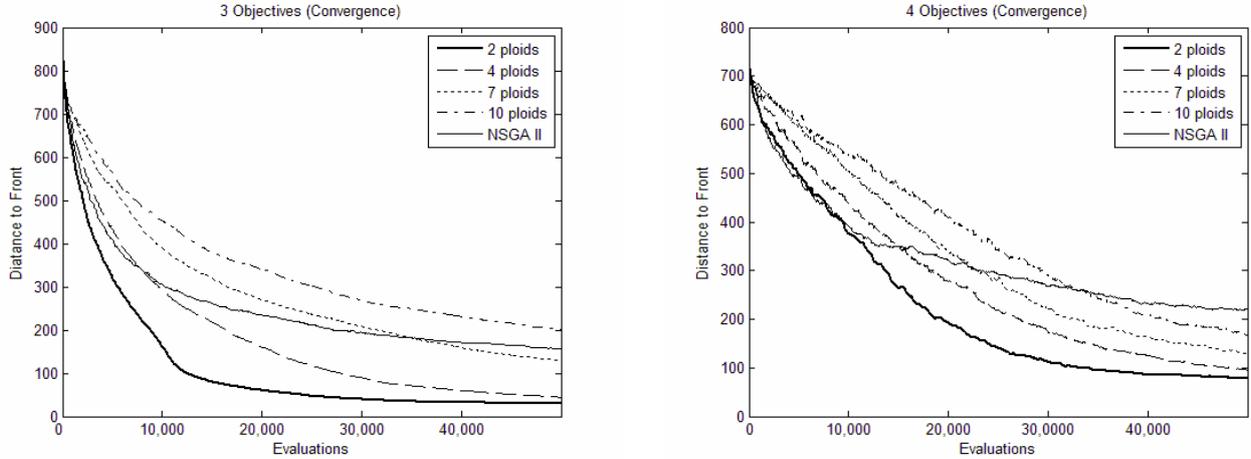

a. 3 objectives      b. 4 objectives

Figure 6 varying ploidy effect on convergence speed for DTLZ1

| MOEA | | Objectives | | | |
|---|---|---|---|---|---|
| | | 3 | 4 | 6 | 10 |
| 2-ploids | Average | 0.6506 | 0.6863 | 0.7067 | 0.7942 |
| | Std. Dev. | 0.0002 | 0.0003 | 0.0002 | 0.0003 |
| 4-ploids | Average | 0.7065 | 0.7075 | 0.7043 | 0.7907 |
| | Std. Dev. | 0.0004 | 0.0001 | 0.0005 | 0.0001 |
| 7-ploids | Average | 0.7124 | 0.7141 | 0.748 | 0.8036 |
| | Std. Dev. | 0.0002 | 0.0002 | 0.0002 | 0.0003 |
| 10-ploids | Average | 0.6993 | 0.6953 | 0.7201 | 0.8198 |
| | Std. Dev. | 0.0003 | 0.0001 | 0.0003 | 0.0001 |
| NSGA-II | Average | 0.6916 | 0.7945 | 0.9312 | 0.8722 |
| | Std. Dev. | 0.0914 | 0.0312 | 0.0195 | 0.0099 |

Table 5 Diversity after 50,000 function evaluations for DTLZ1

| MOEA | Distance to front | Objectives | | | |
|---|---|---|---|---|---|
| | | 3 | 4 | 6 | 10 |
| 2-ploids | Average | 30.37 | 78.38 | 351.08 | 389.06 |
| | Std. Dev. | 0.0785 | 0.0782 | 0.255 | 0.3853 |
| 4-ploids | Average | 44.55 | 95.39 | 418.8 | 382.07 |
| | Std. Dev. | 0.1756 | 0.0828 | 0.3139 | 0.1801 |
| 7-ploids | Average | 127.96 | 129.97 | 447.77 | 426.67 |
| | Std. Dev. | 0.1571 | 0.1657 | 0.1966 | 0.1564 |
| 10-ploids | Average | 201.45 | 167.09 | 466.41 | 414.77 |
| | Std. Dev. | 0.3229 | 0.0905 | 0.2295 | 0.0647 |
| NSGA-II | Average | 157.007 | 219.091 | 442.94 | 485.17 |
| | Std. Dev. | 12.09 | 9.472 | 25.34 | 3.693 |

Table 6 Convergence after 50,000 function evaluations for DTLZ1

Table 5 shows the diversity of the algorithms after 50,000 function evaluations. We can see that the ploids algorithms perform better in the case of 3 objectives, apart from the 2-ploids, while NSGA-II outperforms the other algorithms in the case of 4, 6 and 10 objectives. As in the case of DTLZ2, the diversity of the ploids algorithms is increasing with increasing the number of objectives. A diversity value of 0.65 for the 2-ploids in 3 objectives reaches 0.79 in the 10 objectives case. And finally Table 6 shows the convergence of the algorithms after 50,000 function evaluations. We notice that NSGA-II performs better than the 7 and 10-ploids algorithm in the case of 6 objectives problems. Once again, the 2-ploids algorithm outperforms all the other algorithms in the 3, 4, and 6 objectives. In the 10 objectives case the 4-ploids algorithm is the best, achieving a distance of 382 compared to 389 for the 2-ploids.

## 5.3. DTLZ3

The DTLZ3 test problem is a mix of DTLZ1 and DTLZ2. It has the shape of the later and the huge number of local optima of the former. Again we use the same parameters we used before for recombination and mutation. We use $n = 30$, and go for 50,000 function evaluations. We can see in Figure 7 that once more the NSGA-II is converging well in early evaluations. Then it is overcome by the ploids algorithms one after the other, except the 10-ploids that gets very close to it after 50,000 evaluations. The 2-ploids algorithm achieves best convergence again as seen in Table 8 in the case of 4, 6 and 10 objectives. While the 4-ploids is best for the 3 objectives case, as it reaches a distance to the true Pareto front of 115.5 followed by the 2-ploids with a distance of 141.1. Finally NSGA-II has the worse distribution for the 3 objectives problem with a value of 0.4665, and the second worse is the 10-ploids with 0.5144 distribution value as shown in Table 7. On the other hand, NSGA-II achieves the best diversity values for the 4, 6, and 10 objectives cases. It has a value of 0.7399 in the 6 objectives problem followed by the 10-ploids with a value of 0.5994.

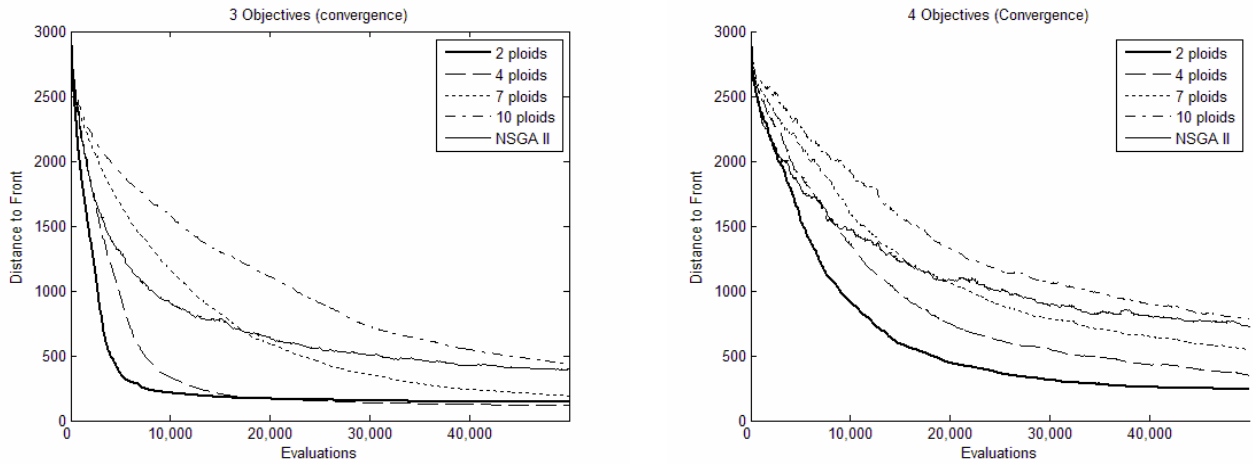

a. 3 objectives  b. 4 objectives

Figure 7 varying ploidy effect on convergence speed for DTLZ3

| MOEA | | Objectives | | | |
|---|---|---|---|---|---|
| | | 3 | 4 | 6 | 10 |
| 2-ploids | Average | 0.6158 | 0.5944 | 0.4768 | 0.6279 |
| | Std. Dev. | 0.0004 | 0.0005 | 0.0004 | 0.001 |
| 4-ploids | Average | 0.638 | 0.5504 | 0.5312 | 0.6736 |
| | Std. Dev. | 0.0005 | 0.0004 | 0.0005 | 0.0008 |
| 7-ploids | Average | 0.5753 | 0.5419 | 0.5554 | 0.7059 |
| | Std. Dev. | 0.0008 | 0.0004 | 0.0006 | 0.0004 |
| 10-ploids | Average | 0.5144 | 0.594 | 0.5994 | 0.6784 |
| | Std. Dev. | 0.0003 | 0.0003 | 0.0006 | 0.0007 |
| NSGA-II | Average | 0.4665 | 0.6217 | 0.7399 | 0.7271 |
| | Std. Dev. | 0.0022 | 0.0002 | 0.0004 | 0.0002 |

Table 7 Diversity after 50,000 function evaluations for DTLZ3

| MOEA | Distance to front | Objectives | | | |
|---|---|---|---|---|---|
| | | 3 | 4 | 6 | 10 |
| 2-ploids | Average | 141.1 | 242.48 | 944.94 | 2363.3 |
| | Std. Dev. | 0.3452 | 0.2461 | 1.4432 | 1.741 |
| 4-ploids | Average | 115.5 | 350.82 | 1137.7 | 2426.3 |
| | Std. Dev. | 0.3571 | 0.4869 | 2.4923 | 1.0542 |
| 7-ploids | Average | 191.24 | 546.1 | 1499.5 | 2393.6 |
| | Std. Dev. | 0.5813 | 0.6889 | 1.742 | 1.1733 |
| 10-ploids | Average | 436.78 | 784.23 | 1669.1 | 2490.1 |
| | Std. Dev. | 0.8684 | 0.9927 | 0.8052 | 1.4455 |
| NSGA-II | Average | 396.2 | 712.24 | 2328.5 | 3124.5 |
| | Std. Dev. | 0.6275 | 0.5085 | 1.4124 | 0.9903 |

Table 8 Convergence after 50,000 function evaluations for DTLZ3

## 5.4. DTLZ4

The final benchmark problem we use is DTLZ4. This problem above anything else tests the ability of the algorithm to keep a diverse set of solutions. The variables are raised to a power of 100, so the algorithm responds by dropping the parameters' values to zero. This leads to a clustered solutions lying on hyper planes with a zero value for the reduced dimension [15]. The ploids algorithms to a good extent overcome this hurdle with reasonably distributed solutions over the Pareto front. We use the same parameters as before with $n = 30$.

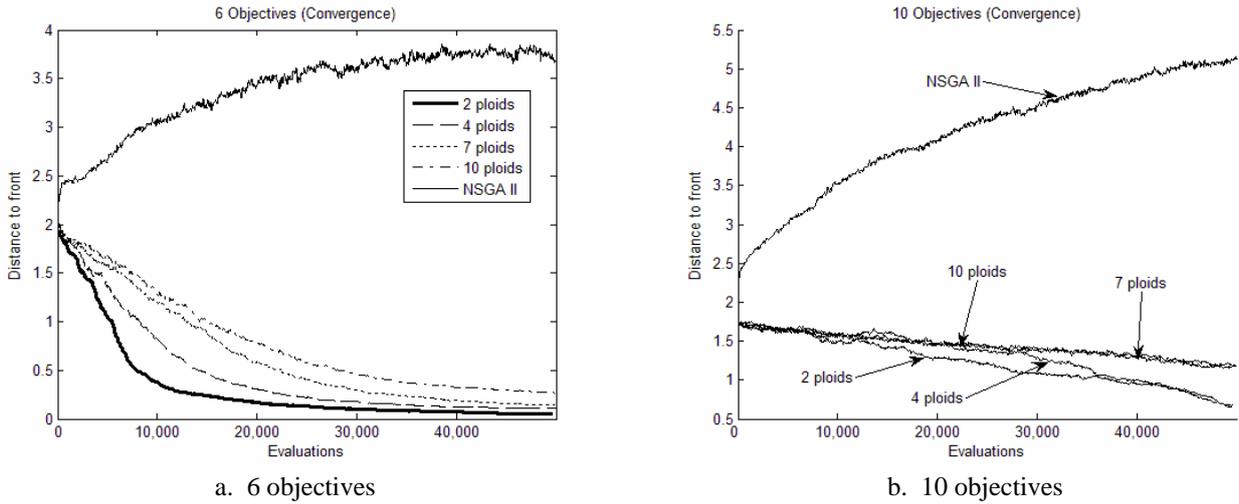

a. 6 objectives  b. 10 objectives

Figure 8 varying ploidy effect on convergence speed for DTLZ4

| MOEA | | Objectives | | | |
|---|---|---|---|---|---|
| | | 3 | 4 | 6 | 10 |
| 2-ploids | Average | 0.5956 | 0.6101 | 0.4946 | 0.1754 |
| | Std. Dev. | 0.001 | 0.0003 | 0.0005 | 0.0005 |
| 4-ploids | Average | 0.6517 | 0.6487 | 0.5634 | 0.1866 |
| | Std. Dev. | 0.0002 | 0.0008 | 0.0002 | 0.0005 |
| 7-ploids | Average | 0.6634 | 0.6926 | 0.5746 | 0.2618 |
| | Std. Dev. | 0.0002 | 0.0002 | 0.0002 | 0.0004 |
| 10-ploids | Average | 0.6711 | 0.6986 | 0.6087 | 0.2763 |
| | Std. Dev. | 0.0003 | 0.0002 | 0.0002 | 0.0004 |
| NSGA-II | Average | 0.732 | 0.687 | 0.697 | 0.7668 |
| | Std. Dev. | 0.0003 | 0.0003 | 0.0002 | 0.0002 |

Table 9 Diversity after 50,000 function evaluations for DTLZ4

| MOEA | Distance to front | Objectives | | | |
|---|---|---|---|---|---|
| | | 3 | 4 | 6 | 10 |
| 2-ploids | Average | 0.0249 | 0.0587 | 0.0529 | 0.6421 |
| | Std. Dev. | 0.0001 | 0.0002 | 0.0001 | 0.0032 |
| 4-ploids | Average | 0.0359 | 0.0547 | 0.1069 | 0.6778 |
| | Std. Dev. | 0.0001 | 0.0001 | 0.0003 | 0.0027 |
| 7-ploids | Average | 0.0748 | 0.0967 | 0.1431 | 1.1748 |
| | Std. Dev. | 0.0002 | 0.0003 | 0.0003 | 0.0012 |
| 10-ploids | Average | 0.2239 | 0.1241 | 0.2654 | 1.1822 |
| | Std. Dev. | 0.0008 | 0.0003 | 0.0008 | 0.0017 |
| NSGA-II | Average | 0.1759 | 0.7957 | 3.7163 | 5.1274 |
| | Std. Dev. | 0.0011 | 0.0032 | 0.0029 | 0.0029 |

Table 10 Convergence after 50,000 function evaluations for DTLZ4

Figure 8 shows the convergence of the algorithms on the 6 and 10 objectives problems. It is interesting to note that in the 10 objectives problem the 4-ploids algorithm outperforms the other algorithms regarding convergence after around 37,000 function evaluations. We notice that NSGA-II totally lost its way in the 6 and 10 objectives problems. The diversity power of the ploids algorithms is apparent in the relatively good distribution for such a hard problem, as shown in Figure 9. For five different runs of the three objectives problem, none of the ploids algorithms clusters on two dimensions losing the dimension of the third, except for a single run for the 2-ploids algorithm. On the other hand, the NSGA-II clusters in three of the five runs, and even in the other two runs it stands on a far distance from the optimal front.

Table 9 presents the diversity of the algorithms on DTLZ4 for different number of objectives. It is clear that the *diversity metric2* fails to detect the clustering of solutions in the case of NSGA-II. It gives the 2-ploids algorithm a diversity value of 0.5956 compared to 0.732 for NSGA-II, though it is clear from Figure 9 that the 2-ploids algorithm has a better diversity. We suggest using a metric that acts on the solutions directly in their $M$-dimensional space (where $M$ is the number of objectives) without projection, such as the S-metric [5]. Finally, Table 10 shows that all the ploids algorithms have a better convergence value than NSGA-II, except for the case of 10-ploids with 3 objectives case. In this case the 10-ploids reaches a distance of 0.2239 to the true Pareto front compared to a value of 0.1759 for NSGA-II. Again the 2-ploids algorithm achieves the best convergence values except for the 4 objectives case where it comes second with a value of 0.0587 to the 4-ploids representation with a value of 0.0547.

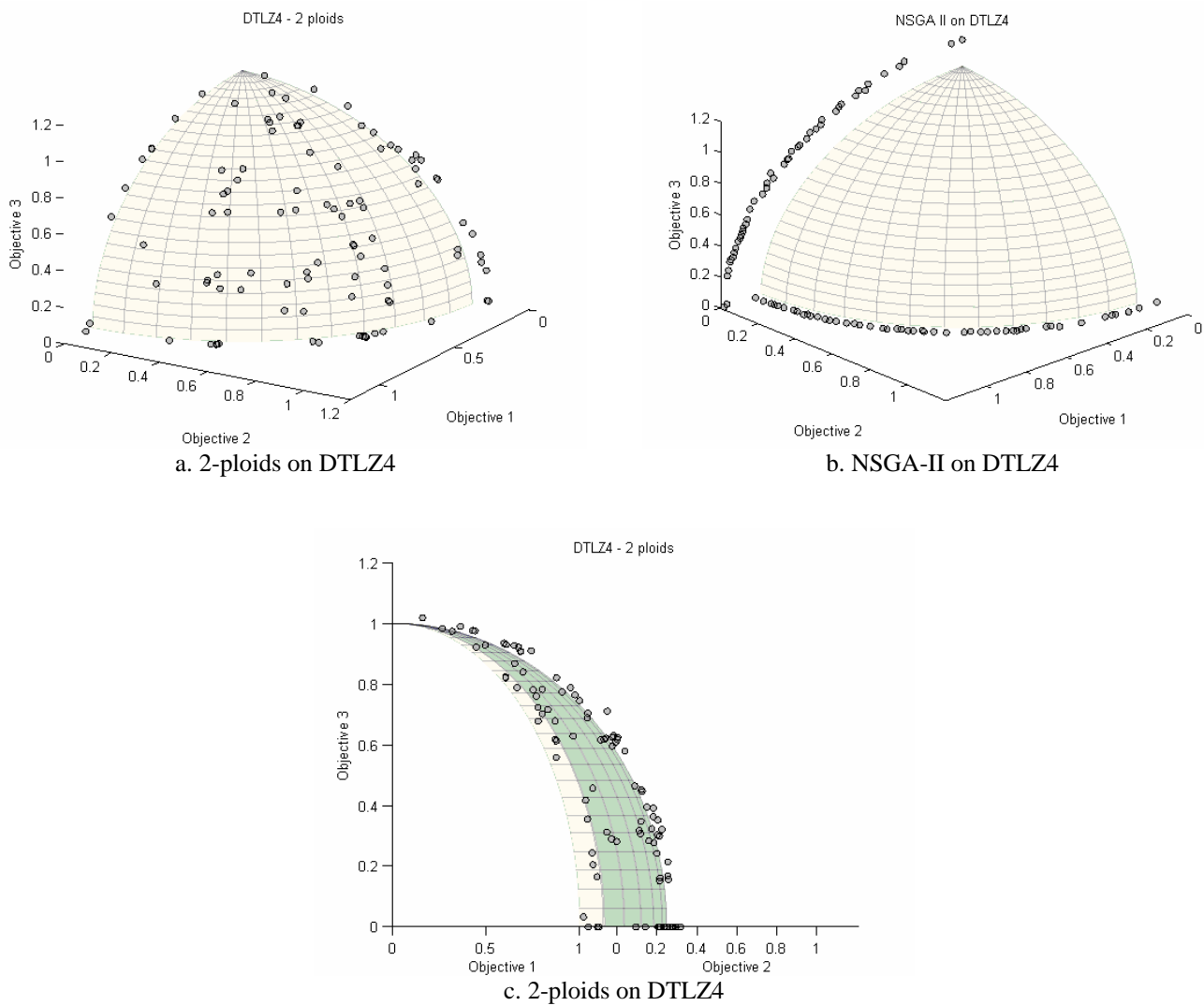

a. 2-ploids on DTLZ4  
b. NSGA-II on DTLZ4  
c. 2-ploids on DTLZ4

**Figure 9**

# 6. Conclusions

The effect of carrying redundant inherited genes to be passed (sometimes silently) through generations causing discontinuous traits inheritance shows promising results for many reasons. First, its relatively quick convergence to the true Pareto front in problems with high number of decision variables and objectives. Second, we can get extra solutions by extracting the redundant chromosomes without any extra evaluations. Third, its ability to maintain a reasonably diverse set of solutions in problems where maintaining good or even reasonable diversity is very difficult. We note that the ploids algorithms achieved good diversity despite the absence of a special diversity maintenance procedure in the algorithm, unlike NSGA-II.

Further investigation is needed. The effect of self adaptation and mutation for the ploids number needs to be analyzed. Autopolyploids which are polyploids with chromosomes derived from a single species, along with allopolyploids which are polyploids with chromosomes derived from different species [6] could be imitated in EAs. The best ploids number for a given problem needs to be determined based on the problem characteristics, or to be made adaptive.